\documentclass[letterpaper, 10 pt, conference]{ieeeconf}  % Comment this line out if you need a4paper
\IEEEoverridecommandlockouts                              % This command is only needed if
\usepackage{cite}
\newtheorem{myDef}{Definition}

\usepackage{amsmath,amssymb,amsfonts}
\usepackage{algorithmic}
\usepackage{graphicx}
\usepackage{subfigure}
\usepackage{textcomp}
\usepackage{xcolor}
\usepackage{color}
\usepackage{siunitx}
\begin{document}

\title{\LARGE \bf A Rule-Compliance Path Planner for Lane-Merge Scenarios Based on Responsibility-Sensitive Safety}

\author{Pengfei Lin$^{1}$, Ehsan Javanmardi$^{1}$, Yuze Jiang$^{1}$, and Manabu Tsukada$^{1}$
% stops a space
\thanks{$^{1}$P. Lin, E. Javanmardi, Y. Jiang, and M. Tsukada are with the Dept. of Creative Informatics, The University of Tokyo, Tokyo 113-8657, Japan. (e-mail: \{linpengfei0609, ejavanmardi, uiryuu, mtsukada\}@g.ecc.u-tokyo.ac.jp)}
}

\maketitle

\begin{abstract}
Lane merging is one of the critical tasks for self-driving cars, and how to perform lane-merge maneuvers effectively and safely has become one of the important standards in measuring the capability of autonomous driving systems. However, due to the ambiguity in driving intentions and right-of-way issues, the lane merging process in autonomous driving remains deficient in terms of maintaining or ceding the right-of-way and attributing liability, which could result in protracted durations for merging and problems such as trajectory oscillation. Hence, we present a rule-compliance path planner (RCPP) for lane-merge scenarios, which initially employs the extended responsibility-sensitive safety (RSS) to elucidate the right-of-way, followed by the potential field-based sigmoid planner for path generation. In the simulation, we have validated the efficacy of the proposed algorithm. The algorithm demonstrated superior performance over previous approaches in aspects such as merging time (Saved 72.3\%), path length (reduced 53.4\%), and eliminating the trajectory oscillation.
\end{abstract}

% \begin{IEEEkeywords}
% Autonomous vehicle, potential field, occlusion, path planning, collision avoidance
% \end{IEEEkeywords}

\section{Introduction}\label{intro}

Autonomous driving technology, utilizing advanced algorithms, sensors, and machine learning, aims to enhance transportation safety and efficiency with reduced human intervention. This shift could significantly lower traffic fatalities by reducing human error. Yet, integrating AVs into current infrastructures presents notable challenges in ensuring algorithm reliability and establishing legal frameworks for AV operation \cite{Teng2023-jj}.

Lane merging is crucial for AV operation, requiring advanced decision-making in diverse traffic scenarios \cite{Shawky2020-cc, Zhu2022-gq}. This involves integrating into an adjacent lane amidst dynamic and unpredictable road conditions, facing challenges such as interpreting traffic and driver intentions while maintaining safety and traffic flow. Recent studies focus on utilizing artificial intelligence and vehicle communications to heighten AVs' prediction and awareness during complex merging, promoting safer, more autonomous transport \cite{Bevly2016-th}.

\begin{figure*}[t]
    \centering
    \includegraphics[width=\hsize]{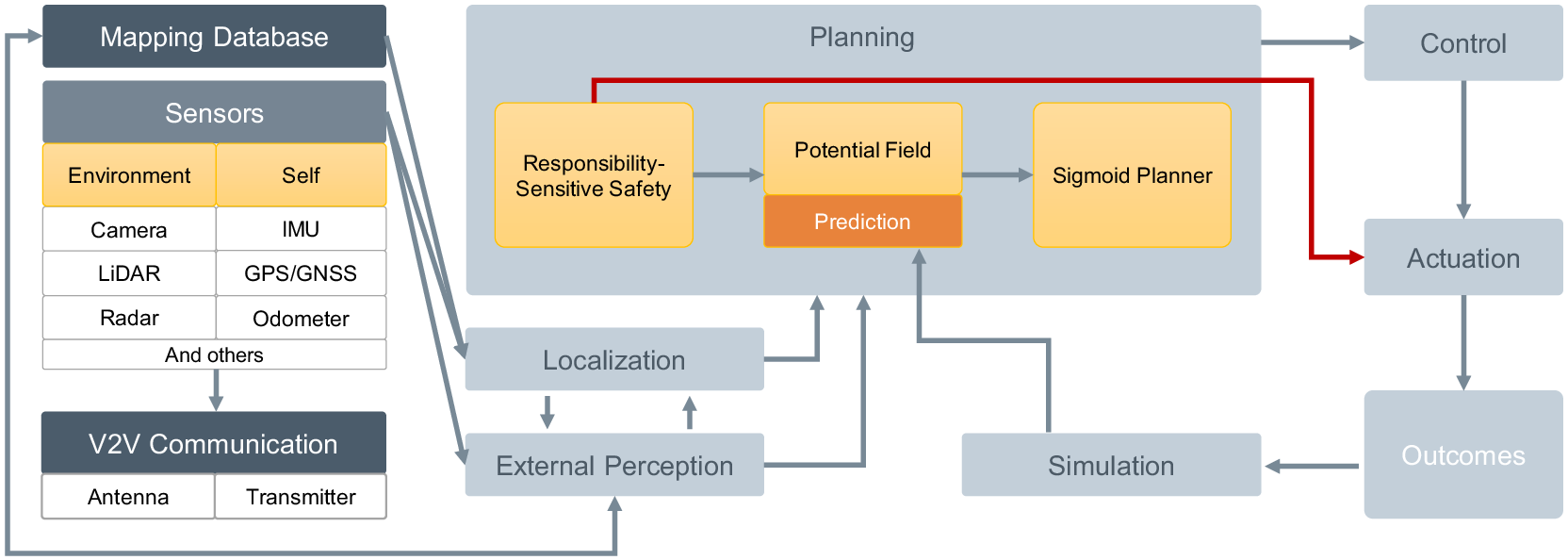}
    \caption{Overall system framework with RCPP: the mapping database and sensors can send raw data to the vehicle-to-vehicle (V2V) communication, external perception, and localization. After that, the featured data is sent to the planning layer, which includes the RSS, the PF, and the sigmoid planner. The reddish-brown solid line indicates that the RSS can enforce commands if emergencies are sensed. Then, the control layer will deliver the commands to the actuation.}
    \label{system_scheme}
\end{figure*}

However, current autonomous lane-merge algorithms encounter issues like trajectory oscillation and ambiguous right-of-way due to limited adaptation and prediction capabilities. Challenges in interpreting human behavior and legal variances further restrict reliable solution development. Hence, we propose a rule-compliance path planning framework to address part of these issues, promoting clear right-of-way and cooperative planning, as depicted in Fig. \ref{system_scheme}. The contributions of this paper are briefly outlined below. 
\begin{itemize}
    \item An extended Responsibility-Sensitive Safety (RSS) model is proposed for clarifying the right-of-way priority and desired collaborative commands.
    \item A potential field (PF)--based sigmoid planner is introduced for lane-merge path generation that can eliminate the amplitude of trajectory oscillations. 
\end{itemize}
The remainder of this paper is outlined as follows: Section II offers a review of related work on lane merging. Section III delineates a comprehensive description of the proposed RCPP method. Comparative simulation results are subsequently described in Section IV, followed by the conclusion in Section V.

\section{Related Work}\label{related_work}

In this section, we review some related works on lane merging associated with autonomous driving. Two principal research paradigms underpin the development of merging algorithms: optimization-based and learning-based approaches. Hu et al. \cite{Hu2019-eb} introduced an online system control algorithm for CAV multilane freeway merging, focusing on optimizing lane change and car following trajectories to balance traffic flow. This method integrates a Cooperative Lane Changing Control (CLCC) model and a Cooperative Merging Control (CMC) model, coordinated via a dynamic border point approach. Then, Jing et al. \cite{Jing2019-yh} presented a cooperative multi-player game-based optimization framework to globally and optimally coordinate CAVs in merging zones, aiming to minimize fuel consumption, improve passenger comfort, and reduce travel time. Subsequently, Xu et al. \cite{Xu2019-pv} introduced a grouping-based cooperative driving strategy as a compromise, fixing the passing order for vehicles with small enough headways to reduce the solution space and improve efficiency. Liu et al. \cite{Liu2021-gm} developed the Leader-Follower Game Controller (LFGC) for AVs, addressing forced merge scenarios through a partially observable leader-follower game model. Utilizing Model Predictive Control (MPC), the LFGC predicts other vehicles' intentions and trajectories, enhancing safety and efficiency in merging. Zhu et al. \cite{Zhu2022-fr} presented a flow-level CAV coordination strategy for multilane freeway merging, incorporating lane-change rules, proactive gap creation, and vehicle platooning to enhance traffic flow and efficiency. Recently, Ji et al. \cite{Ji2023-gj} developed TriPField, a new three-dimensional potential field model, combining an ellipsoidal potential field with a Gaussian velocity field to refine path-planning for AVs in dynamic, dense settings. This model surpasses traditional isotropic PFs in computational efficiency and accuracy in modeling multi-vehicle interactions. However, optimization-based lane merging in autonomous driving struggles with computational complexity, limited scalability in unpredictable environments, and dependency on extensive, often inaccessible, information, hindering real-world applicability and coordination.

Recently, Chen et al. \cite{Chen2023-oa} presented a multi-agent reinforcement learning (MARL) framework that accommodates dynamic, time-varying traffic and promotes inter-agent cooperation through parameter sharing and local rewards, while employing action masking and a priority-based safety supervisor to enhance learning efficiency and safety. And then, Arbabi et al. \cite{Arbabi2023-ls} introduced a decision-making approach for autonomous driving, targeting the challenge of merging into moving traffic under uncertainty from other drivers and imperfect sensors. They formulate this as a partially observable Markov decision process (POMDP) and employ a Monte Carlo tree search to devise a strategy for complex maneuvers, integrating a predictive model of traffic dynamics and interactions. Later on, Ammourah et al. \cite{Ammourah2023-nw}, Gu et al. \cite{Gu2023-co}, Guo et al. \cite{Guo2023-wk}, and Gurses et al. \cite{Gurses2023-yu} all focus on enhancing autonomous driving strategies through RL approaches. They explore the efficient development of driving strategies, cooperative control between traffic infrastructure and AVs, safety enhancements, and specific maneuvers such as mandatory lane changing. However, RL-based lane merging methods primarily suffer from the need for extensive training data, vulnerability to dynamic and unpredictable environments, and difficulty in ensuring safety due to the exploratory nature of the algorithms. Therefore, employing rules to assist with lane merging tasks, akin to human drivers who adhere to traffic regulations, the introduction of rule-guided approaches is becoming a trend.

\section{Rule-Compliance Path Planning}\label{pf}

This section introduces the proposed rule-compliance path planning for lane merging scenarios, including the extended RSS, PF, and sigmoid planner.

\subsection{Extended Responsibility-Sensitive Safety for Merging}
\begin{figure*}[t]
    \centering
    \includegraphics[width=\hsize]{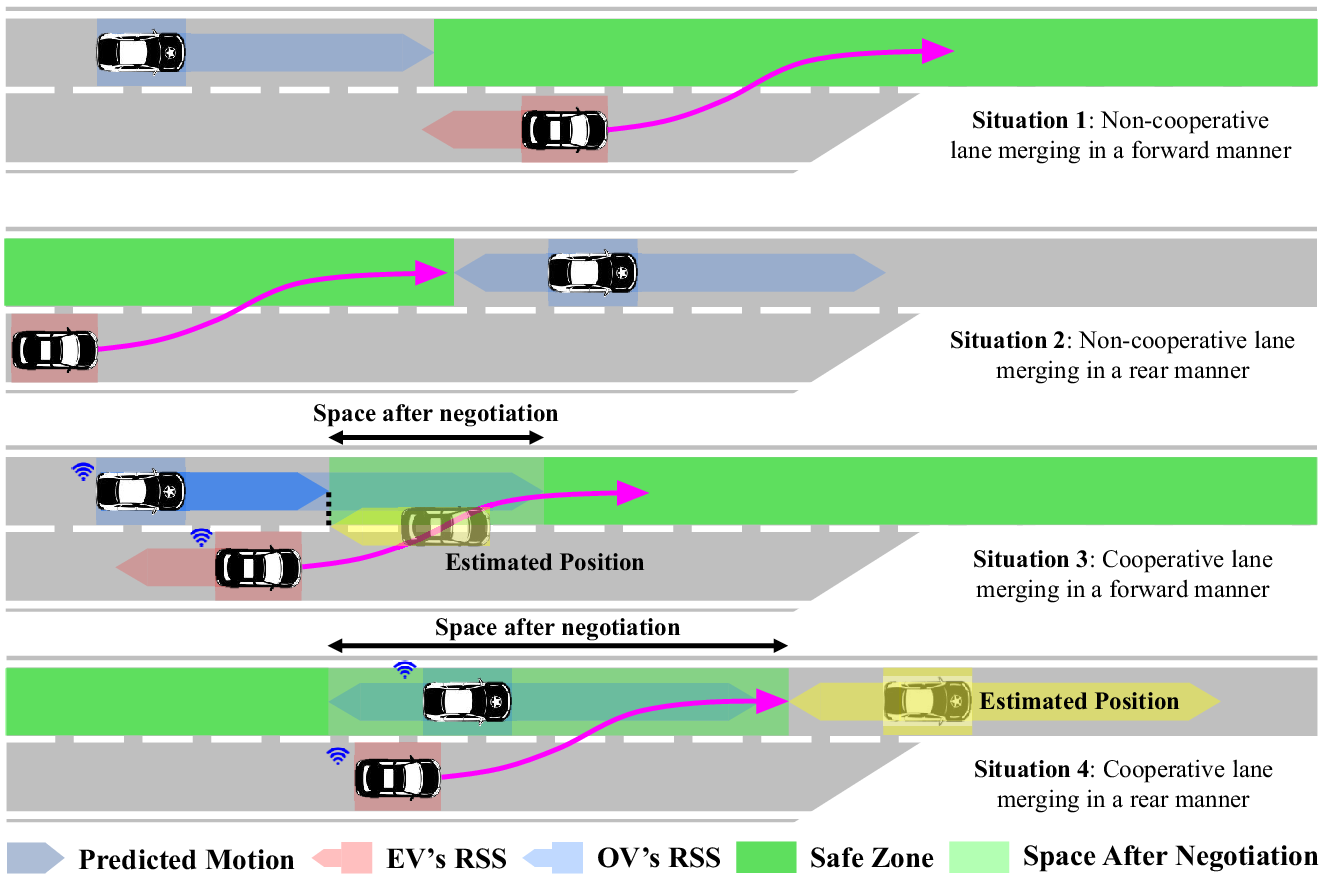}
    \caption{Merging situations with different traffic conditions}
    \label{merge_explanation}
\end{figure*}

The RSS proposed by Intel/Mobileye is a safety model designed for autonomous driving systems \cite{Shalev-Shwartz2017-bh, Gassmann2019-ou}. It provides a formal framework aimed at ensuring that self-driving cars operate safely and avoid causing accidents. The fundamental principle of the RSS involves establishing clear, definitive rules that govern autonomous vehicle behavior across various traffic situations to avert collisions, emphasizing the maintenance of safe inter-vehicle distances, adherence to right-of-way, and execution of secure lane changes and merges. Nevertheless, while the RSS framework establishes foundational standards for assorted geometric road scenarios, the particulars concerning lane merging behaviors and right-of-way allocation remain indistinct, especially on highways. Moreover, existing rules often lead to overly conservative driving behaviors as they primarily consider extreme situations \cite{Liu2021-uc}. Therefore, we propose an extended version of the RSS tailored to safe but faster lane merging behaviors under both non-cooperative and cooperative driving situations. First, we introduce two essential concepts of the RSS: minimum lateral and longitudinal safe distances $D_{rss}^{lat, i}$ and $D_{rss}^{long, i}$ \cite{Shalev-Shwartz2017-bh}.
\begin{align}
    \begin{split} 
        D_{rss}^{long,i}=
        &[ v_eT_{lag}+\frac{1}{2}a_{accel}^{max}T_{lag}^2\\
        &+\frac{l_e+l_f^i}{2}+\frac{(v_e+ a_{accel}^{max}T_{lag})^2}{2a_{brake}^{min}}-\frac{{v_f^i}^2}{2a_{brake}^{max,i}} ]_+,
        \label{long_rss}
    \end{split}\\
    \begin{split}
        D_{rss}^{lat,i}=
        &\mu+\Bigg[\frac{v_e^{lat}+v_{e,\rho}^{lat}}{2}T_{lag}+\frac{{v_{e,\rho}^{lat}}^2}{2a_{brake}^{lat,min}}+\frac{l_w+l_{w}^i}{2}\\
        &-\left(\frac{v_{o}^{lat,i}+v_{o,\rho}^{lat,i}}{2}T_{lag}+\frac{{v_{o,T}^{lat,i}}^2}{2a_{brake}^{lat,min,i}}\right)\Bigg]_+,
        \label{lateral_rss}
    \end{split}
\end{align}
where $v_e$ and $v_f^i$ are the longitudinal speeds of the ego and front vehicles, respectively. $T_{lag}$ represents the retardation time, including the response time of the ego vehicle and the communication delay between Connected AVs (CAVs). $a_{accel}^{max}$ and $a_{brake}^{min}$ correspond to the ego vehicle's maximum acceleration capacity and minimum deceleration threshold, respectively. Additionally, $a_{brake}^{max, i}$ indicates the maximum braking capability of the $i^{th}$ obstacle vehicle. The variables $l_e$ and $l_f^i$ are utilized to denote the lengths of the ego vehicle and the $i^{th}$ obstacle vehicle, respectively. $\mu$ determines the fluctuation margin and the superscript $lat$ denotes the lateral information. $l_w$ and $l_w^i$ are the widths of the ego and $i^{th}$ obstacle vehicles, respectively. Next, we introduce the rule-compliance lane merging approach, predicated on minimum safe distances, within non-cooperative and cooperative driving contexts. We extend two definitions based on the current RSS regulations for different merging situations that could arise on the highway, as exampled in Fig. \ref{merge_explanation}. 
Herein, the first definition is provided to regulate scenarios depicted in the non-cooperative lane merge (Situation 1 and 2 in Fig. \ref{merge_explanation}), encompassing appropriate responses and contingency plans.
\begin{myDef}[\textbf{Non-cooperative Lane Merge}] Let $c_o^j$ represent the $i^{th}$ obstacle vehicle on the main lane, and let $c_e$ denote the ego vehicle on the side lane merging into the main lane. Let $\rho_{m}$ be the lane-merge threshold time, and let $T_{m}^{dec}$ signify the lane-merge decision time. A merging maneuver is deemed safe for the two non-cooperative vehicles traveling in the same direction, provided they comply with the following constraints:

    \textbf{1.} If at the interval $[T_{m}^{dec},\;T_{m}^{dec}+\frac{\rho_{m}}{2}]$, and say that $c_e$ is ahead $c_o^i$; then:
    \begin{equation}
        X_{o}^i+\frac{v_o^i\rho_{m}}{2}+\frac{a_{accel}^{max,i}\rho_{m}^2}{8}\leq X+v_e^*\frac{\rho_{m}}{2}-D_{rss}^{long}
        \label{ineq_1}
    \end{equation}
    
    \textbf{2.} If at the interval $[T_{m}^{dec},\;T_{m}^{dec}+\rho_{m}]$, and say that $c_e$ is behind $c_o^i$; then:
    \begin{equation}
        X+v_e^*\rho_{m}\leq X_{o}^i+v_o^i\rho_{m}-D_{rss}^{long,i}
        \label{ineq_2}
    \end{equation}
    
    \textbf{3.} If continuous insufficient space prevents $c_e$ from merging, then $c_e$ must halt entirely before reaching the end of the side lane, pausing for an opportunity to re-merge.
\label{def1}
% \vspace{-0.66cm}
\end{myDef}

In Definition \ref{def1}, the first constraint presupposes that the following vehicle $c_o^j$ may accelerate up to $a_{accel}^{max,j}$ until $T_{LC}^{dec}+\frac{\rho_{LC}}{2}$, then adapts its speed as per RSS when $c_e$ merges. The second constraint mandates that $c_e$ maintains its speed based on empirical guidance during merging, adjusting according to the main lane's traffic velocities $v_o^{j-1}$ and $v_o^{j}$ before the lane change. $X$ and $X_o^i$ are the longitudinal position of the ego vehicle and $i^{th}$ obstacle vehicle, respectively. Besides, cooperative driving is essential when merging space does not satisfy Definition \ref{def1} criteria. For human drivers, this involves signal usage; for CAVs, it entails V2V communication to share necessary collaborative data. Rules for this process are outlined in Fig. \ref{merge_explanation} (Situation 3 and 4), as shown below.
\begin{myDef}[\textbf{Cooperative Lane Merge}] Let $c_e$, $c_o^i$, and $T_{m}^{dec}$ be as defined in Definition \ref{def1}, with $D_{rss}^{long,*}$ indicating $c_e$'s predicted minimum safe distance at the sigmoid CP, and $\rho_{c}$ as the communication threshold.  A merging maneuver is deemed safe if these vehicles comply with the outlined responses:

    \textbf{1.} If at the interval $[T_{m}^{dec},\;T_{m}^{dec}+\rho_{c}]$, and say that $c_e$ is ahead $c_o^i$; then:
    \begin{enumerate}
        \item[A.] $c_e$ must send $c_o^i$ a standard V2V message with $P_c$ and $D_{rss}^{long,*}$ included.
        \item[B.] $c_o^i$ must brake at most $a_{brake}^{min,i}$ until reaching $c_o^{i,*}$ with the given constraint:
        \begin{small}
            \begin{equation}   
                X_o^i+\frac{v_o^i+v_o^{i,*}}{2}\rho_{c}+v_o^{i,*}\frac{\rho_{m}}{2}\!\leq\! P_c-D_{rss}^{long,*}
                \label{ineq_3}% X+v_e\rho_{CT}+
            \end{equation}
        \end{small} 
    \end{enumerate}

    \textbf{2.} If at the interval $[T_{m}^{dec},\;T_{m}^{dec}+\rho_{c}]$, and say that $c_e$ is behind $c_o^i$; then:
    \begin{enumerate}
        \item[A.] $c_e$ must send $c_o^i$ a standard V2V message, and then brake at most $a_{brake}^{min}$
        \item[B.] $c_o^i$ must accelerate at most $a_{accel}^{max,i}$ until reach $v_o^{i,*}$ with the given constraint:
        \begin{small}
            \begin{equation}   
                X+v_e\rho_{c}-\frac{a_{brake}^{min}{\rho_{c}}^2}{2}\leq X_o^i+\frac{v_o^{i,*}+v_o^i}{2}\rho_{c}-D_{rss}^{long,i}
                \label{ineq_4}
            \end{equation}
        \end{small}
    \end{enumerate}
    
    \textbf{3.} If at the interval $[T_{m}^{dec},\;T_{m}^{dec}+\rho_{c}]$, and say that $c_o^i$ is non-cooperative or non-responsive to $c_e$; then $c_e$ must merge by following Definition \ref{def1}.
\label{def2}
\end{myDef}

Notably, in non-cooperative situations, the ego vehicle's speed threshold $v_e^*$ is determined by solving inequalities \ref{ineq_1} and \ref{ineq_2}. In cooperative situations, it's assumed the ego vehicle seeks collaboration at current speed $v_e$, thus determining the desired cooperative speed $v_o^{i,*}$ for the obstacle vehicle based on \ref{ineq_3} and \ref{ineq_4}. If the obstacle vehicle $c_o^j$ remains non-cooperative or unresponsive, the ego vehicle defaults to non-cooperative merging as per Definition \ref{def1}.

\subsection{Potential Field}

The potential field (PF) is used to conceptualize the vehicle's environment as a field of forces, where obstacles exert repulsive forces and goals exert attractive forces on the vehicle. In autonomous driving, the PF is utilized for modeling and analyzing road traffic environments, which assigns repulsive forces to road markings, such as road boundary lines and lane lines, thereby ensuring that vehicles remain within their designated lanes. The subsequent equation is employed to characterize the intended behavior of the road markings \cite{Wolf2008-ye}.
\begin{equation}
    P_{rm}=\frac{1}{2}\beta \left(\frac{1}{Y-Y_{l,r}-\frac{1}{2}l_w}\right)^2,
    \label{road_marks}
\end{equation}
where $\beta$ denotes the repulsive coefficient of the road's potential field. $Y_{l,r}$ denotes the positions of the left and right boundaries of the road, respectively, in the global coordinate frame. $l_w$ indicates the vehicle's width, considering the vehicle's dimensions. Similarly, the PF establishes repulsive fields on obstacle vehicles to facilitate obstacle avoidance via the following equation \cite{Ji2017-fk}.
\begin{equation}
    P_{ob}=\gamma\left|exp\left(-\left[\sigma_1\left(Y-Y_o\right)^2+\sigma_2\left(X-X_o\right)^2\right]\right)-U\right|,
    \label{obstacle}
\end{equation}
where $\gamma$ denotes the influence coefficient pertaining to the obstacle's potential field. $\sigma_{1,2}$ represent the lateral and longitudinal coefficients of the repulsive field, respectively, which is computed by Eqs. (\ref{long_rss}) and (\ref{lateral_rss}). $U$ is defined as a minimal positive factor. Finally, the lane center produces an attractive field to draw the vehicle towards it. However, this can be overridden if a lane merge is initiated, as the formula below implements. 
\begin{equation}
    P_{lc}=
    \begin{cases}
        \xi d(X_{d},Y_{d})^2/2,&{\text{if lane keeping}}\\
        D^*\xi d(X_{d},Y_{d})-\xi \left(D^*\right)^2/2,&{\text{if lane merging}},
    \end{cases}
    \label{lane_center}
\end{equation}
where $\xi$ is the coefficient influencing the attractive potential, $d(X_{d}, Y_{d})$ denotes the distance between the vehicle and the target waypoint sampled from the lane center, and $D^*$ signifies the designated search target distance.

\subsection{Sigmoid Planner}
\begin{figure}[t]
    \centering
    \includegraphics[width=\hsize]{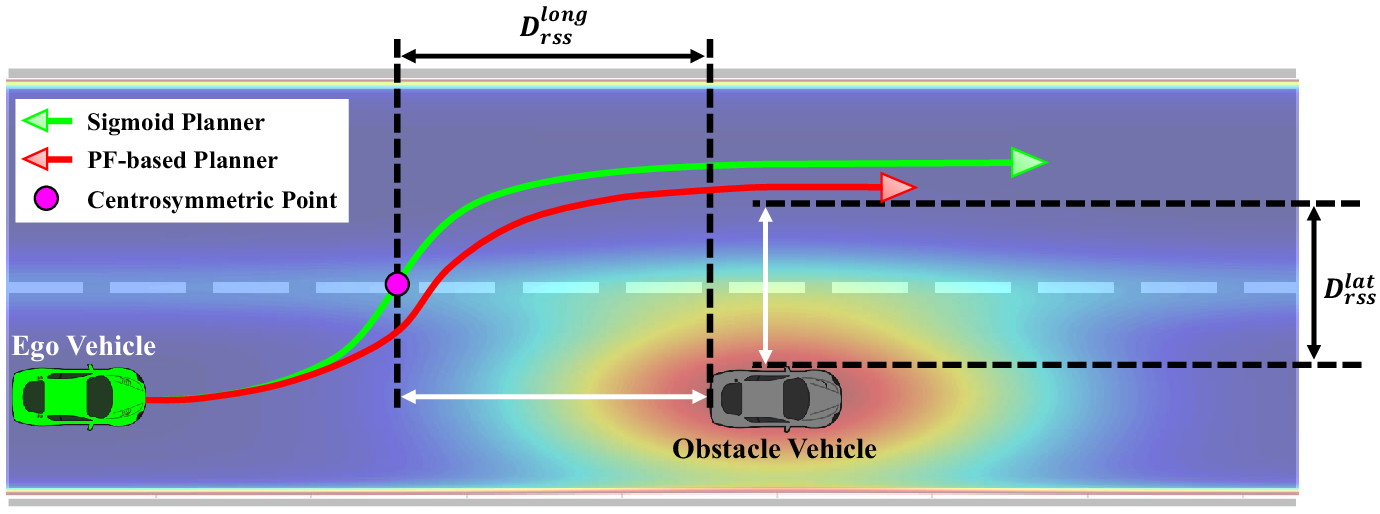}
    \caption{Proposed sigmoid planner based on the PF under the RSS criteria}
    \label{sigmoid_planner}
\end{figure}
\begin{figure*}[t]
    \centering
    \includegraphics[width=\hsize]{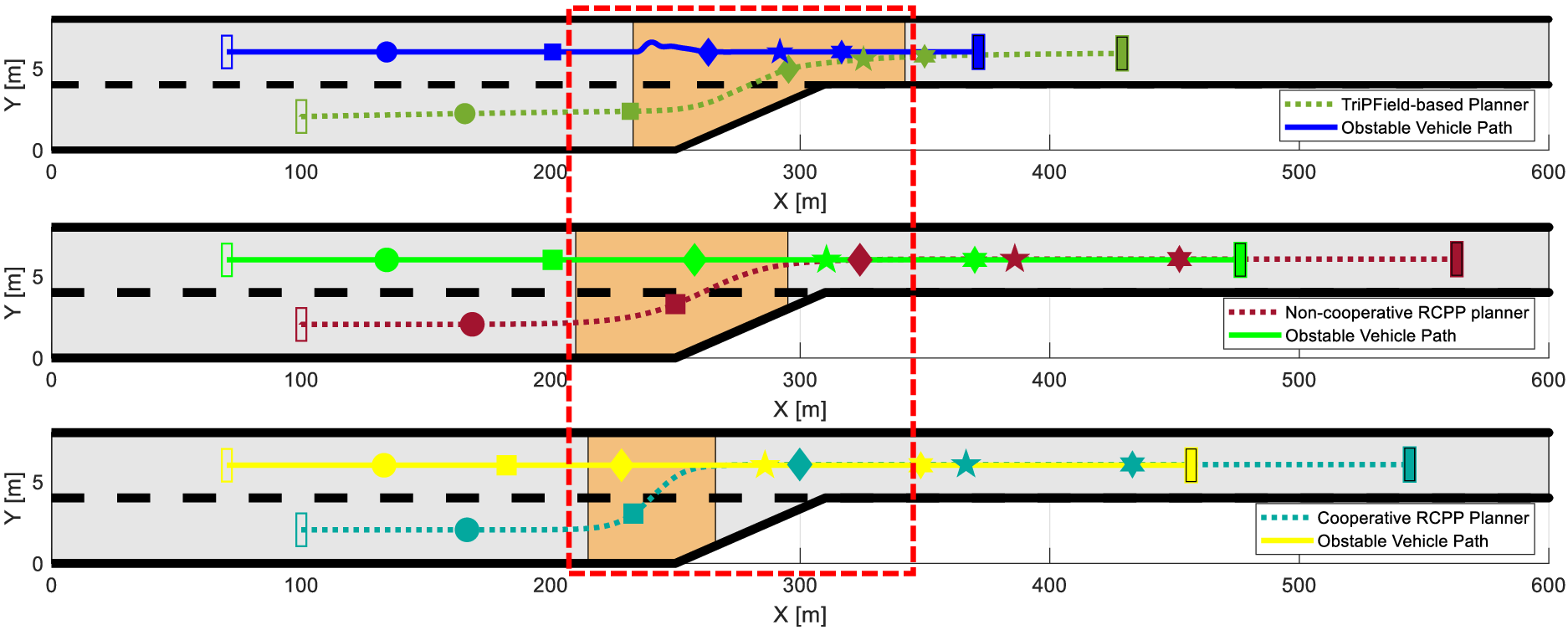}
    \caption{Lane-merge paths with TriPField-based Planner, Non-cooperative RCPP Planner, and Cooperative Planner}
    \label{2D_paths}
\end{figure*}

Although the PF is widely used due to its simple structure and high real-time performance, some studies indicate that PF-based path planning encounters local minima issues, affecting the quality of local path generation \cite{Ji2023-gj, Lin2023-ja}. Consequently, to ensure a more stable lane-change path, a sigmoid curve known for producing easily navigable paths is proposed \cite{Cesari2017-jx}. Differing from the nonlinear optimization approach proposed by Lu et al. \cite{Lu2020-kk}, we use the PF to construct the sigmoid curve according to the RSS criteria, prioritizing safety-critical considerations and time efficiency. The canonical representation of the sigmoid function, $f_{sig}$, is specified as follows.
\begin{equation}
    f_{sig}(X)=
    \frac{W}{1+e^{-\kappa(X-P_c)}}+b,
\end{equation}
where $W$ denotes the straight-line distance between the centers of adjacent lanes, $\kappa$ signifies the slope at the midpoint of the sigmoid curve; $P_c$ denotes the longitudinal position of the midpoint, and $b$ represents the offset from the X-axis. As depicted in Fig. \ref{sigmoid_planner}, the computation of $P_c$ typically involves two different constraints based on the road conditions.
\begin{equation}
    \begin{cases}
        P_c \leq X_o-D_{rss}^{long},&{i=1}\\
        X_o^i+D_{rss}^{long,*}\leq P_c\leq X_o^{i+1}-D_{rss}^{long,i+1},&{i\geq 1}
    \end{cases}
\end{equation}
It should be noted that the above inequality can be untenable if $X_o^i+D_{rss}^{long,*}\geq X_o^{j+1}-D_{rss}^{long,j+1}$. Under such circumstances, the ego vehicle should abort the lane-merge maneuver and react appropriately to the prevailing conditions.

\section{Simulation Results}\label{AA}

This section delineates the settings and outcomes of the simulation analysis. Furthermore, this study encompasses a comparative analysis involving three lane-merge planners: (i) the TriPField-based method \cite{Ji2023-gj}; (ii) the proposed RCPP method with non-cooperative; and (iii) the proposed RCPP method with cooperative. We set the initial positions and starting velocities of the ego and obstacle vehicles relatively close to each other to simulate lane merging scenarios under ambiguous conditions.

As shown in Fig. \ref{2D_paths},  The ego vehicle with TriPField-based Planner (depicted by the dashed green line) merges into the lane from an entry point that appears around X=226 m. The path towards merging (highlighted area) shows significant deviation and weaving, indicating attempts to adjust speed and position to find an optimal merging spot. The blue line represents the obstacle vehicle path, showing that the ego vehicle successfully merges behind the obstacle vehicle. The merging appears to be smooth as indicated by the path becoming parallel to the obstacle vehicle's path beyond the merging zone. In contrast, the ego vehicle with the non-cooperative RCPP planner (brown dotted line) merges into the lane with less deviation compared to the TriPField-based Planner, indicating a more straightforward approach to merging. However, the close proximity to the obstacle vehicle suggests a potentially aggressive or less cautious merging strategy, which might not be ideal in heavy traffic or uncertain conditions. Lastly, the dashed yellow line indicates the ego vehicle that uses a Cooperative RCPP planning approach. This path (cyan dotted line) shows a smooth transition into the lane with minimal deviations, suggesting effective coordination between the ego vehicle and the obstacle vehicle. The merging path aligns well with the movements of the obstacle vehicle, indicating a high level of cooperation and communication, likely resulting in a safer and more efficient merge.
\begin{figure*}
\centering
    \subfigure[Sideslip angle]{
        \begin{minipage}[t]{0.49\linewidth}
            \centering
            \includegraphics[width=\hsize]{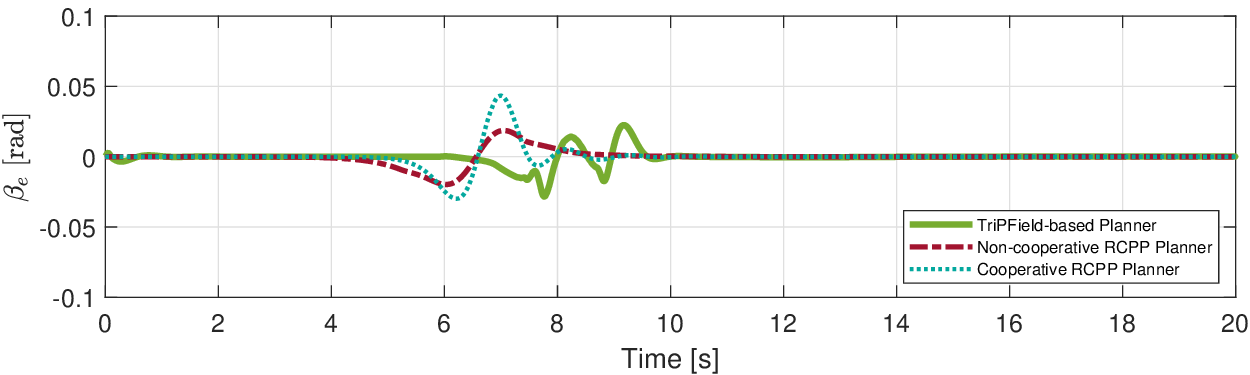}
        \end{minipage}
    \label{beta}
    }%
    \subfigure[Yaw angle]{
        \begin{minipage}[t]{0.49\linewidth}
            \centering
            \includegraphics[width=\hsize]{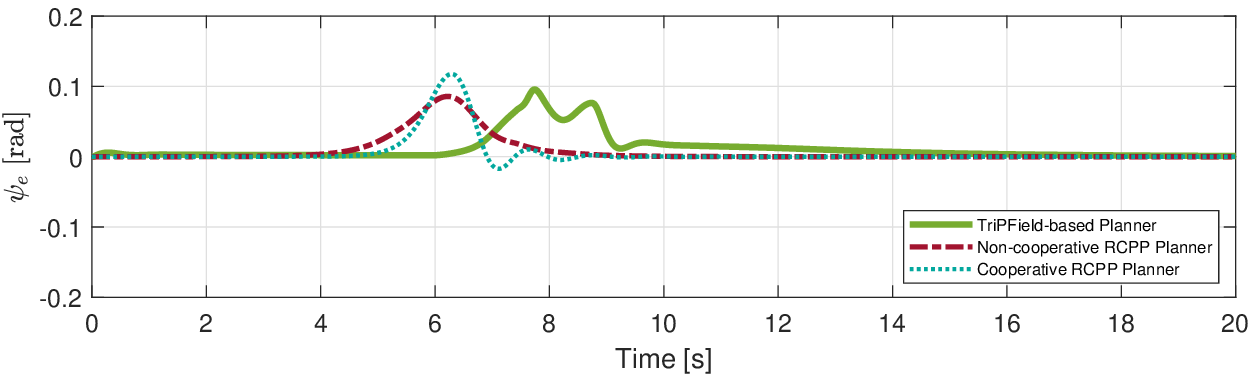}
        \end{minipage}
    \label{psi}
    }
    
    \subfigure[Longitudinal speed of the obstacle vehicle]{
        \begin{minipage}[t]{0.49\linewidth}
            \centering
            \includegraphics[width=\hsize]{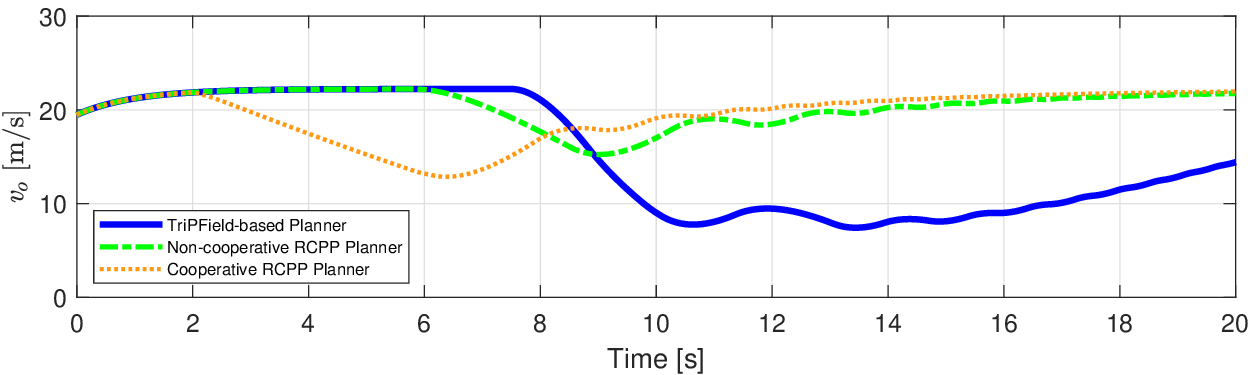}
        \end{minipage}
    \label{v_obs}
    }%
    \subfigure[Longitudinal speed of the ego vehicle]{
        \begin{minipage}[t]{0.49\linewidth}
            \centering
            \includegraphics[width=\hsize]{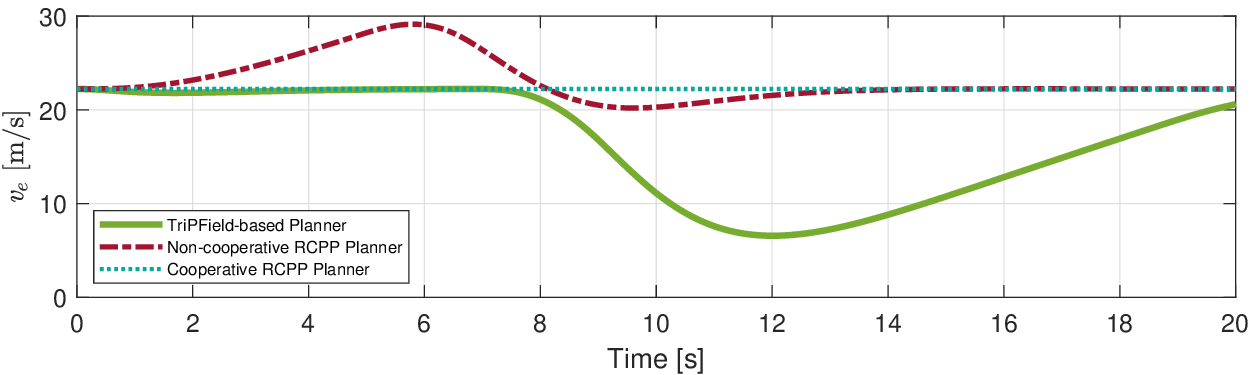}
        \end{minipage}
    \label{v_ego}
    }
\caption{Motion states of the ego and obstacle vehicles}
\label{motions}
\end{figure*}

The motion states are illustrated in Fig. \ref{motions}. Fig. \ref{beta} shows the sideslip angle, where the TriPField-based planner (green solid line) oscillates between 7.6 s and 9.7 s with an amplitude below $\pm$0.02 rad. In contrast, the non-cooperative RCPP planner (brown chain line) maintains a steadier sideslip between -0.012 rad and 0.016 rad, whereas the cooperative RCPP planner (red dashed line) peaks at 0.024 rad at 7.1 s, indicating a faster lane merge but at the cost of a larger sideslip angle. Fig. \ref{psi} reveals that the TriPField planner exhibits aggressive turning, while both RCPP planners show more stable, controlled turning, with minor vibrations in the cooperative approach. In Fig. \ref{v_obs}, the obstacle vehicle using the TriPField-based planner decelerates sharply below 10 m/s, whereas the non-cooperative RCPP planner decelerates more gradually to around 16 m/s. The cooperative RCPP planner decelerates early, providing space for the ego vehicle's merging after receiving a cooperation message. In Fig. \ref{v_ego}, the TriPField planner decelerates sharply to under 9 m/s, the non-cooperative RCPP planner briefly accelerates to 30 m/s then slows to 20 m/s post-merging, while the cooperative RCPP planner maintains a steady speed without acceleration changes.

\section{Conclusion}

In this study, we presented a Rule-Compliance Path Planner (RCPP) for autonomous driving lane-merge scenarios, which incorporates an extended Responsibility-Sensitive Safety (RSS) framework to address right-of-way ambiguities and employs a PF-based sigmoid planner for trajectory generation. Our simulations demonstrate the RCPP's significant improvements in merging time, path length, and trajectory stability over traditional methods. By enhancing safety, efficiency, and rule compliance in lane merging, our approach advances autonomous driving technologies toward more predictable and harmonious integration into current traffic systems.

% \section*{Acknowledgment}

% These research results were obtained from the commissioned research by the National Institute of Information and Communications Technology (NICT), JAPAN, and the Japan Society for the Promotion of Science (JSPS) KAKENHI (grant number: 21H03423), and partly sponsored by the China Scholarship Council (CSC) program (No.202208050036) and JSPS DC program (grant number: 23KJ0391).

% %
% \begin{figure}[t]
%     \centering
%     \includegraphics[width=\hsize]{figs/alpha_new.eps}
%     \caption{$\alpha_1$ and $\alpha_2$ from PF-OAPP}
%     \label{alpha}
% \end{figure}
% %
% \begin{figure}[t]
%     \centering
%     \includegraphics[width=\hsize]{figs/occlusion_flag.eps}
%     \caption{Occlusion flag: 1 means the occlusion appears; 0 means the occlusion disappears}
%     \label{occlu_flag}
% \end{figure}
% \end{thebibliography}
\bibliographystyle{IEEEtran}
\bibliography{Reference.bib}

\end{document}